\documentclass[10pt,twocolumn,letterpaper]{article}

\usepackage[utf8]{inputenc}
\usepackage[T1]{fontenc}
\usepackage[final,algorithms]{wacv}

\newif\ifsubmissioncopy
\submissioncopytrue     

\usepackage{siunitx}
\DeclareSIUnit{\frameunit}{frame}

\ifsubmissioncopy

    \newcommand{\todo}[1]{}
    \newcommand{\TODO}[1]{}
    \newcommand{\gb}[1]{}
    \newcommand{\yl}[1]{}
    \newcommand{\lock}[1]{}
    \newcommand{\ad}[1]{}
    \newcommand{\rd}[1]{}
    \newcommand{\rp}[1]{}

\else

    \newcommand{\todo}[1]{\textcolor{red}{#1}}

    \newcommand{\TODO}[1]{\textbf{\textcolor{red}{[TODO: #1]}}}
    \newcommand{\gb}[1]{\textcolor{red}{[GB: #1]}}
    \newcommand{\yl}[1]{\textcolor{violet}{[YL: #1]}}
    \newcommand{\lock}[1]{\textcolor{red}{[Locked and editing locally -- do not edit on Overleaf: #1]}}
    \newcommand{\ad}[1]{\textcolor{cyan}{[AD: #1]}}
    \newcommand{\rd}[1]{\textcolor{purple}{[RD: #1]}}
    \newcommand{\rp}[1]{\textcolor{brown}{[RP: #1]}}

\fi

\usepackage{wrapfig}
\usepackage{array}
\usepackage{longtable}
\usepackage{multirow}
\usepackage{nicefrac}
\usepackage{microtype}
\usepackage{placeins}

\usepackage[normalem]{ulem} 

\definecolor{wacvblue}{rgb}{0.21,0.49,0.74}
\usepackage[pagebackref,breaklinks,colorlinks,allcolors=wacvblue]{hyperref}
\usepackage[capitalize]{cleveref}

\usepackage[accsupp]{axessibility}  

\AtBeginDocument{%
  \crefname{figure}{Fig.}{Figs.}%
  \Crefname{figure}{Figure}{Figures}%
  \crefname{table}{Table}{Tables}%
  \Crefname{table}{Table}{Tables}%
  \crefname{section}{Sec.}{Secs.}%
  \Crefname{section}{Section}{Sections}%
  \crefname{appendix}{Appendix}{Appendices}%
  \Crefname{appendix}{Appendix}{Appendices}%
  \crefname{equation}{Eq.}{Eqs.}%
  \Crefname{equation}{Equation}{Equations}%
}

\newcommand{\latin}[1]{\textit{#1}\xspace} 

\renewcommand{\ie}{\latin{i.e.}}
\renewcommand{\etal}{\latin{et al.}}

\newcommand{\alg}{eSharp\xspace}

\def\wacvPaperID{57}
\def\confName{WACV}
\def\confYear{2027}

\title{\alg: Diffusion-based Sharp Monocular Metric Event-to-Depth Estimation}

\author{
Qiushi Dai$^{1}$\thanks{Equal contribution.}\quad
Aniket Dashpute$^{1*}$\quad
Yuhao Liu$^{1}$\quad
Rishubh Parihar$^{2}$\\
Guha Balakrishnan$^{1}$\quad
Ashok Veeraraghavan$^{1}$\thanks{Corresponding author: \texttt{vashok@rice.edu}.}\\[2pt]
$^1$Rice University \qquad $^2$Indian Institute of Science
}

\begin{document}
\maketitle

\begingroup
\renewcommand\thefootnote{}
\footnotetext{Accepted to the IEEE/CVF Winter Conference on Applications of Computer Vision (WACV) 2027.}
\addtocounter{footnote}{0}
\endgroup

\begin{abstract}

Depth perception is fundamental for autonomous navigation, industrial inspection, and mixed-reality systems. 
It often operates under strict constraints on power and form factor. In such cases, monocular depth estimation (MDE) is often used. Even as recent advances continue to improve monocular depth quality, conventional frame-based MDE systems remain challenged by fast motion and high-dynamic-range scenes. 
This work introduces \alg, an MDE framework that addresses these problems using an event camera. 
\alg repurposes pretrained diffusion priors for event-only MDE and produces accurate metric depths with unparalleled sharpness on real data. 
In contrast, existing event-based MDE methods often fail to recover fine spatial detail. 
Real-world event datasets have only sparse LiDAR depth maps available for ground truth supervision, making dense prediction challenging. We address this with cross-modal distillation, using LiDAR-calibrated dense pseudo-labels generated from a spatially co-registered frame camera to train \alg. 
Experiments show that \alg outperforms prior event-to-depth baselines in both metric depth accuracy and spatial fidelity, establishing diffusion models as a promising foundation for event-driven 3D perception
\footnotemark{}\footnotetext{Source code available at \url{https://github.com/rice-computational-imaging/eSharp}}.
\end{abstract}
\vspace{-10pt}

\section{Introduction}
\vspace{-1mm}
\label{sec:introduction}

\begin{figure}[h]
    \centering
    \includegraphics[width=\columnwidth]{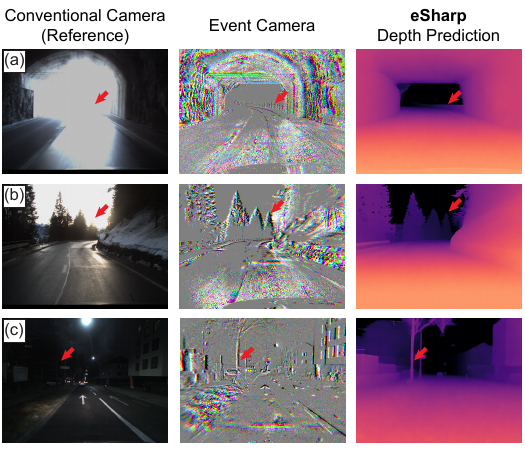}
    \vspace{-7mm}
    \caption{\textbf{eSharp produces sharp, accurate depth maps under challenging conditions.} Under imaging constraints such as (a) high dynamic range, (b) sensor saturation, and (c) low light, conventional frame cameras cannot provide visual signals for reliable depth estimation. Event cameras are better suited for these challenging imaging conditions, and this work proposes \alg, a dense metric depth estimator conditioned on monocular event cameras.}
    \label{fig:intro}
    \vspace{-6mm}
\end{figure}

Recent advances in 3D imaging and multimodal fusion have enabled many complex tasks, such as autonomous navigation~\cite{liu2023bevfusion,cui2024survey}, augmented/mixed reality~\cite{liu2025efficient,wang2023practical} and medical imaging applications~\cite{wang2022neural,hayoz2024online}.
Many applications, such as drone navigation, are further constrained by utility, form factor, and hardware complexity. In such settings, monocular (single-camera) depth estimation (MDE) is particularly attractive because it requires minimal sensing hardware, leading to rapid progress in this area~\cite{wang2025vggt,yang2024depth,ke2024repurposing}.

Current MDE methods, while powerful, rely on conventional frame cameras, which suffer in applications involving fast motion or high-dynamic-range scenes.
These cameras integrate intensity over finite exposure windows and require a minimum readout time per frame.
This acquisition process limits temporal resolution and the dynamic range of the cameras.
Rapid motion and extreme changes in illumination can thus cause loss of edges, texture, and contrast, leading to inaccurate or incomplete depth predictions.
In autonomous driving, for example, a rapid transition from a tunnel into daylight (\cref{fig:intro}a) or strong back-lighting (\cref{fig:intro}b) can saturate bright regions, whereas objects in low-light scenes may be severely underexposed and visually indistinguishable (\cref{fig:intro}c).
In contrast, \textit{event cameras}~\cite{gallego2020event} are neuromorphic differential sensors well-suited to these scenarios.
They asynchronously detect per-pixel changes in log intensity rather than capturing complete image frames.
Their differential sensing mechanism and sparse readout allow high temporal resolution and wide dynamic range (60–80 dB more than frame cameras), making them apt for depth estimation under rapid motion and challenging illumination conditions~\cite{gehrig2024low}.

Despite these sensing advantages, MDE from events alone remains challenging. 
Event cameras' sparse readout
provides no absolute intensity measurements and can miss regions with insufficient texture.
Previous work has therefore estimated depth by fusing events with conventional intensity frames in stereo~\cite{mostafavi2021event,ghosh2025event} or monocular~\cite{gehrig2021combining} systems, which provide complementary cues for depth perception.
More recent methods perform event-only monocular depth estimation~\cite{e2depth,depthanyevent,lee2025distile2d}, eliminating the need for a synchronized frame stream but making depth inference more under-constrained, as scene geometry must be recovered from sparse event measurements.
Consequently, the existing event-only monocular depth estimators often produce over-smoothed predictions with poorly localized depth discontinuities on real data.
Moreover, evaluation is typically limited to sparse LiDAR-valid pixels, leaving large regions and fine structures unevaluated. Aggregate metrics, therefore, do not fully capture the completeness or geometric fidelity of predicted depth.

To address these challenges, we propose adapting a pretrained diffusion model as the backbone of our monocular depth estimator with dense pseudo-depth ground truth to enable the generation of sharp images.
First, the dense metric depth labels for training are generated by a novel, accurate process using a frame-to-depth teacher MDE model~\cite{yang2024depth}.
Second, we leverage the diffusion model's strong pretrained priors over images and depth maps to produce sharp, semantically consistent estimates.
In practice, we fine-tune Marigold~\cite{ke2024repurposing} and Lotus~\cite{he2025lotus} as the event-to-depth backbones.
To adapt to this event-sensing domain, we fine-tune them by replacing the RGB image input with a three-channel event representation. This is formed by partitioning a time window into three sub-windows and simply accumulating events within each.
Third, these priors learned from large-scale natural scenes and depth maps enable strong generalization beyond the event-camera training distribution.
We call our proposed event-based MDE framework \textit{\alg}~\footnote{Proposed model \alg is pronounced as the musical note \textit{E$\sharp$}.}.

In summary, event cameras overcome the sensing limits imposed by frame cameras, while diffusion models and dense ground-truth labels bridge the gap in monocular depth estimation quality between frame and event modalities.

\noindent
Our main contributions are as follows:
\begin{itemize}    
    \item \textbf{Structure-preserving} sharp, high-accuracy metric depth estimates from monocular event streams, also showing a downstream segmentation task on the depth maps.
    \item Novel \textbf{dense pseudo depth} based training setup for event-to-depth estimation
    \item Direct \textbf{metric} depth estimation that outperforms prior methods even with their LiDAR-based scale alignment.
    \item \textbf{Zero-shot generalization} for event to depth prediction for out-of-distribution scenes
\end{itemize}

\begin{figure*}[h]
    \centering
    \includegraphics[width=\textwidth]{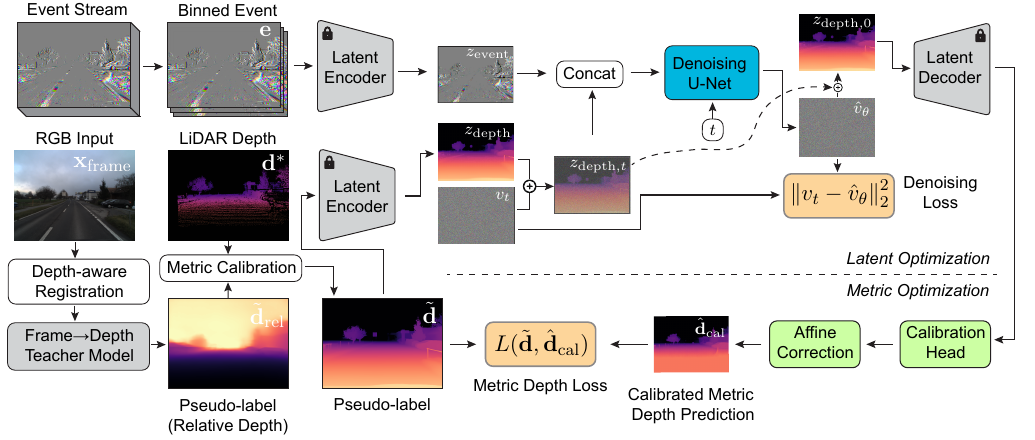}
    \vspace{-6mm}
    \caption{\textbf{Overview of \alg.}
    The event stream is temporally binned and encoded into an event latent.
    An RGB-to-depth teacher generates dense pseudo-depth supervision, which is calibrated to LiDAR and encoded to depth latent.
    We concatenate the clean event latent and the noisy depth latent.
    Training proceeds in two stages: Stage I adapts the denoising U-Net for event-conditioned depth generation, while Stage II jointly fine-tunes the denoising U-Net and a lightweight calibration head to improve metric depth predictions.}
    \label{fig:framework}
    \vspace{-4mm}
\end{figure*}
\section{Related Work}
\vspace{-1mm}
\label{sec:related-work}

\noindent
\textbf{Monocular depth estimation}. Monocular depth estimation is a difficult inverse problem that involves generating depth maps from a single-camera image. Initial papers in this field used a multi-scale deep network~\cite{eigen2014depth}, which was followed by Monodepth~\cite{godard2017unsupervised} using a convolutional network, Monodepth2~\cite{godard2019digging} using U-Net architectures.
MiDaS~\cite{ranftl2020towards} introduced large datasets and showed zero-shot cross-dataset transfer, and DPT~\cite{ranftl2021vision} introduced a transformer-based architecture for this task.
Marigold~\cite{ke2024repurposing} introduced sharp depth maps using a pretrained latent diffusion model. This branch of models was further developed in CH$_3$Depth~\cite{li2025ch} and Lotus~\cite{he2025lotus}.
More recent vision transformer (ViT) based models, such as DepthAnything/v2/v3~\cite{yang2024depthv1,yang2024depth,lin2025depth}, Depth Pro~\cite{bochkovskiy2025depth}, Metric3D~\cite{yin2023metric3d}, VGGT~\cite{wang2025vggt}, and MoGe 1,2,3~\cite{wang2025moge,wang2026moge,kong2026fine}, have further pushed the state of the art for metric MDE.

\vspace{5pt}

\noindent
\textbf{Event-based depth estimation.}
Previous work in event-based depth estimation has often used either a stereo system of event cameras or an event-frame pair~\cite{ghosh2025event}. 
Zhu~\etal~\cite{zhu2019unsupervised} first predicted sparse monocular depth jointly with relative pose using a feed-forward neural network.
Dense monocular event depth estimation was first proposed in e2depth~\cite{e2depth} using a recurrent architecture.
EMoDepth~\cite{zhu2023self} uses a U-Net-style architecture with cross-modal consistency from event-aligned intensity images during training for test-time monocular estimation.
EReFormer~\cite{ereformer} proposes the use of a recurrent transformer modeling both the global spatial context and rich temporal cues from the event stream.
More recently, EDE-Distill~\cite{hu2025ede}, DepthAnyEvent~\cite{depthanyevent}, EventDAM~\cite{zhu2025depthanyeventstream}, and Distil-E2D~\cite{lee2025distile2d} use cross-modal distillation to transfer spatial and semantic knowledge from an image-based teacher to an event-based student model.
AIMDepth~\cite{jing2026aimdepth} and ESDepth~\cite{wang2026esdepth} build on state space models and perform well on the real dataset MVSEC~\cite{mvsec}, but do not show results on the more recent diverse DSEC~\cite{dsec}.
More recently, GLU-Net uses uncertainty optimization with a global-local event fusion network ~\cite{wu2026glu}.
RAMNet~\cite{gehrig2021combining}, SRFNet~\cite{pan2024srfnet}, FUSE~\cite{sun2025fuse}, UniCT~\cite{jing2025unict}, and the full model in HypoDepth~\cite{liu2026depth} perform joint image-event MDE, requiring a frame image as an additional input and thus operating in a different sensing setting from event-only methods.

Recent event-only MDE methods have made substantial progress, but dense predictions on real-world datasets such as MVSEC~\cite{mvsec} and DSEC~\cite{dsec} often lack the spatial detail seen on simulated datasets such as EventScape~\cite{gehrig2021combining} and DENSE~\cite{e2depth}. In addition, the strongest reported estimates are often obtained by scaling relative-depth predictions with LiDAR measurements rather than by directly predicting metric depth. Motivated by these gaps, \alg produces sharp and accurate metric depth on real-world data, preserves scene structure for downstream segmentation, and qualitatively generalizes to out-of-domain data such as TUM-VIE~\cite{tumvie}.

\section{Preliminaries}
\label{sec:preliminaries}

\noindent \textbf{Diffusion models} \citep{song2020scorebased,ho2020denoising,rombach2022highresolution} are generative models that generate images by learning to iteratively remove noise from a complete Gaussian noise. Latent diffusion models~\cite{rombach2022highresolution}
perform denoising in the latent space of a pretrained
VAE~\cite{kingma2014autoencoding} for efficiency. Furthermore, these models~\cite{rombach2022highresolution} can be trained on text to perform text conditioned image generation. In a variance-preserving noise schedule~\citep{ho2020denoising}, given a clean latent $z_0$, the forward process produces:

\vspace{-5mm}
\begin{equation}
\label{eq:ddpm}
z_t = \alpha_t z_0 + \sigma_t \epsilon,
\quad
v_t = \alpha_t \epsilon - \sigma_t z_0,
\quad
\epsilon \sim \mathcal{N}(0,I),
\end{equation}

\vspace{-1mm}

\noindent 
where $\alpha_t$ and $\sigma_t$ are signal and noise coefficients, and $v_t$ is the $v$-prediction (velocity) target. A denoising network $v_\theta$ predicts $v_t$ from the noisy latent $z_t$, the conditioning input, and timestep $t$. An alternative parameterization is $x_0$-prediction, in which the denoiser directly estimates the clean latent $z_0$ rather than the velocity. This formulation reduces prediction variance in dense prediction tasks and is adopted by Lotus~\citep{he2025lotus}.
Although $x_0$ prediction does not inherently require single-step inference, Lotus couples it with training at only the terminal timestep $t=T$, allowing the denoiser to estimate $z_0$ in a single network evaluation.
We adopt this formulation for a faster version of our method (\alg-fast), which therefore uses one denoising step at inference.

\smallskip \noindent \textbf{Monocular depth estimation} aims to infer a depth map $\mathbf{d} \in \mathbb{R}_{+}^{H \times W}$ from a single image $\mathbf{x} \in \mathbb{R}^{H \times W \times C}$, where channel $C = 3$ for RGB images.
Given the excellent ability of large diffusion models in generating realistic images, Marigold~\citep{ke2024repurposing} repurposed the Stable Diffusion model to perform monocular frame-to-depth estimation.
Specifically, it concatenates the image latent with a noisy depth latent at each denoising step, and fine-tunes the denoising U-Net to recover the depth latent, which is decoded by the frozen VAE decoder to produce a dense depth map. Our proposed event-to-depth model follows the same latent depth-generation principle as Marigold, but works with much sparser encoded event stream rather than dense RGB input image.

\smallskip \noindent \textbf{Event cameras} respond to per-pixel changes in log intensity rather than capturing images at a fixed frame rate. When the change in log intensity at a pixel exceeds a set threshold $\epsilon$ (controlled through bias parameters of the camera), an event is generated and read out asynchronously.
An $i$-th event is represented as $e_i=(x_i,y_i,t_i,p_i)$, where $(x_i,y_i)$ is the pixel location, $t_i$ is the timestamp, and $p_i\in\{-1,+1\}$ is the polarity of the intensity change.
The resulting sequence of sparse events forms an event stream.
The differential sensing and sparse asynchronous readout of event cameras enable high dynamic range and low latency.

\vspace{-1mm}
\section{Method}
\label{sec:method}
\vspace{-1mm}

Event-only MDE is typically trained using events accumulated over a temporal window together with a corresponding depth map for supervision. In real-world datasets such as DSEC~\cite{dsec} and MVSEC~\cite{mvsec}, LiDAR depth measurements are spatially and temporally aligned with the event stream and commonly serve as ground-truth depth. 
These datasets additionally provide temporally aligned frame images for reference. In MVSEC, the frames are directly from the event cameras while for DSEC, they are from a separate camera.
We train \alg exclusively on real-world event data because existing simulators~\citep{dosovitskiy2017carla} do not fully capture the event-generation process and sensor nonidealities of physical event cameras. As a practical example, event cameras operate in near-continuous time, whereas simulated events are derived from upsampled discrete frames, limiting their actual temporal precision. An overview of \alg is shown in \cref{fig:framework}; its components are detailed in the following subsections.

\subsection{Dense Metric Pseudo-Depth Supervision.}
\label{sec:pseudo_label}

As LiDAR sensors provide ground-truth depth measurements only at sparsely sampled points, getting dense supervision for real data is challenging. Similar to recent works \cite{depthanyevent,lee2025distile2d}, we address this gap using cross-modal distillation: during training, we use temporally aligned frame images and co-register them to the event-camera image plane to generate pseudo-depth labels.
We distill depth priors from a frame-to-depth model to train an event-to-depth model.

For each temporal window of the event stream, we obtain the corresponding frame image $\mathbf{x}_{\mathrm{frame}}$ from the neighboring frame camera in the dataset. We first geometrically register this image to the event-camera image plane (for DSEC).
We use Depth Anything V2~\cite{yang2024depth} to predict dense relative depth and the frame-camera intrinsics to back-project it into a relative 3D point cloud $\mathbf{x}_{\mathrm{3D-rel}}$.
We then use the frame-aligned LiDAR depth map to recover metric scale, yielding $\mathbf{x}_{\mathrm{3D}}$.
Using the inter-camera extrinsics, we transform these 3D points into the event-camera coordinate system and project them onto the event-camera image plane, yielding the registered frame $\mathbf{x}^{\mathrm{e}}_{\mathrm{frame}}$.
In contrast to approximate frame-event registration used in prior work~\cite{gehrig2024low,depthanyevent}, this accurate depth-aware registration accounts for scene geometry.

Next, we run the same MDE teacher model, Depth Anything V2~\cite{yang2024depth} on $\mathbf{x}^{\mathrm{e}}_{\mathrm{frame}}$ to obtain a normalized dense relative-depth map $\tilde{\mathbf{d}}_{\mathrm{rel}}\in[0,1]^{H\times W}$.
We then convert $\tilde{\mathbf{d}}_{\mathrm{rel}}$ to metric depth $\tilde{\mathbf{d}}$ by least-squares fitting over pixels with valid ground-truth LiDAR measurements $\mathbf{d}^{\ast}$ (\ie, sparse ground-truth depth).
For datasets with low-resolution grayscale frames, the frame-to-depth teacher becomes less reliable, so we apply quantile matching instead of least-squares fitting to improve pseudo-label robustness; details are provided in Supplementary  \cref{sec:supp_pseudo_label_generation}.

The resulting pseudo-labels combine dense RGB frame-derived scene structure with metric LiDAR measurements and are used only during training, requiring neither RGB frames nor LiDAR at inference.

\subsection{Event-Conditioned Latent Diffusion}
\label{sec:event_representation}

\begin{figure}[t]
    \centering
    \includegraphics[width=\columnwidth]{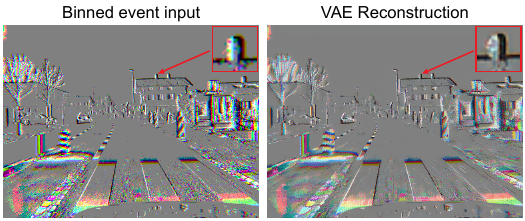}
    \vspace{-6mm}
    \caption{\textbf{VAE preserves binned event semantics.} We feed binned event image $\mathbf{e}$ through a frozen VAE (without denoising). The overall event semantics are well-preserved.}
    \label{fig:vae_compression}
    \vspace{-15pt}
\end{figure}

To condition the pretrained image diffusion model on asynchronous events, we adopt a simple uniform-binning representation.
Unlike prior approaches that use linearly weighted accumulation within uniform bins~\cite{e2depth,depthanyevent} or nonuniform temporal intervals~\cite{nam2022stereo}, we find that equal-duration binning is sufficient for effective conditioning.
Specifically, we uniformly partition the event window $[T_0,T_1]$ into three temporal bins and accumulate the signed event polarities within each bin.
This produces a three-channel event tensor $\mathbf{e}\in\mathbb{R}^{H\times W\times3}$ compatible with the VAE~\cite{kingma2014autoencoding} used in Stable Diffusion~\cite{rombach2022highresolution}.

The VAE maps $\mathbf{e}$ to event latent $z_{\mathrm{event}}\in\mathbb{R}^{h\times w\times4}$ and the dense pseudo-depth $\tilde{\mathbf{d}}$ to depth latent $z_{\mathrm{depth}}\in\mathbb{R}^{h\times w\times4}$.
As seen in \cref{fig:vae_compression}, this VAE largely retains binned event semantics, making it a suitable event encoder even without finetuning.
We follow Marigold~\cite{ke2024repurposing} to repurpose Stable Diffusion's denoising U-Net to condition on two image inputs, shown in \cref{fig:framework}.
At each diffusion timestep, the noisy depth latent and event latent are concatenated along the feature dimension, resulting in an eight-channel input to the denoising U-Net.
Accordingly, we double the input dimensionality of the U-Net's first convolution from four to eight channels.
Consequently, the diffusion denoising process is conditioned on event observations to generate the depth latent.

We introduce two variants of our model based on the pretrained MDE diffusion model we adapt. Our base model \alg is fine-tuned on Marigold~\cite{ke2024repurposing} and operates best in the native 10-step diffusion inference mode. We also have a fast version of this model that does one-step inference which we call \alg-fast. This is fine-tuned on Lotus~\cite{he2025lotus}.

\subsection{Two-Stage Event-to-Depth Adaptation}

Adapting the pretrained diffusion models to event inputs and recovering accurate metric depth involves two objectives: learning a sharp depth map prediction and estimating the correct shift and scale values for accurate metric prediction.
We therefore separate training into two stages: first, we adapt and fine-tune a pretrained MDE diffusion model to recover dense scene structure from event observations; second, we explicitly correct residual global scale and shift errors for metric depth prediction.

\smallskip \noindent \textbf{Event-Conditioned Diffusion Adaptation.}
Given event and depth latents $z_{\mathrm{event}}$ and
$z_{\mathrm{depth}}$, we train each model using its native diffusion
objective. For \alg, we sample timestep $t$ and optimize the
$v$-prediction objective. For \alg-fast, we follow Lotus and directly
predict the clean depth latent from the terminal timestep $T$.
The first and second cases below correspond to \alg and \alg-fast,
respectively:
\vspace{-1mm}
\begin{equation}
\label{eq:diff}
\mathcal{L}_{\mathrm{diff}} =
\begin{cases}
\mathbb{E}\!\left[
\left\|
v_\theta(z_{\mathrm{depth},t},z_{\mathrm{event}},t)-v_t
\right\|_2^2
\right],\\
\mathbb{E}\!\left[
\left\|
x_\theta(z_{\mathrm{depth},T},z_{\mathrm{event}},T)
-z_{\mathrm{depth}}
\right\|_2^2
\right].
\end{cases}
\end{equation}
Both objectives teach the model to recover scene geometry conditioned
on event observations. The predicted clean depth latent is decoded by
the frozen VAE decoder to produce $\hat{\mathbf{d}}$, whose residual depth errors are corrected in the next stage.

\smallskip \noindent \textbf{Metric Depth Calibration.} Despite optimizing the event-conditioned diffusion objective in \cref{eq:diff}, the decoded prediction $\hat{\mathbf{d}}$ can retain sample-dependent global scale and shift errors.
We explicitly address this metric ambiguity via a lightweight calibration head $g_{\phi}$ that estimates a frame-specific scale and shift $(s,\beta)=g_{\phi}(\hat{\mathbf{d}})$ and produces the calibrated metric prediction $\hat{\mathbf{d}}_{\mathrm{cal}}=s\hat{\mathbf{d}}+\beta$.
The calibration head consists of a compact convolutional encoder followed by a two-layer MLP that regresses the two affine parameters.

To supervise the calibration head, we obtain target parameters
$(s^*, \beta^*)$ by least-squares fitting the current prediction
to the dense metric pseudo-depth $\tilde{d}$ over all image
pixels $\mathcal{P}$ (see Supplementary \cref{sec:supp_metric_depth_calibration}).
The calibration objective supervises both the affine parameters
and calibrated metric depth:
\vspace{-1mm}
\begin{equation}
\begin{aligned}
\mathcal{L}_{\mathrm{calib}}
={}&
\lambda_{\mathrm{aff}}
\left[
\operatorname{SmoothL1}(s,s^*)
+
\operatorname{SmoothL1}(\beta,\beta^*)
\right]
\\
&+
\frac{\lambda_{\mathrm{metric}}}{|\mathcal{P}|}
\sum_{u \in \mathcal{P}}
\left|
\hat{d}_{\mathrm{cal}}(u)-\tilde{d}(u)
\right|.
\end{aligned}
\label{eq:calib}
\end{equation}

\vspace{-1mm}

The SmoothL1 term provides robust supervision for the predicted scale and shift, while the second term directly penalizes metric depth error after calibration (see Supplementary for details).
During Stage II, the diffusion U-Net and calibration head are jointly optimized:

\vspace{-0.5mm}
\begin{equation}
    \label{eq:stage2}
    \mathcal{L}_{\mathrm{stage\,II}}
    =
    \mathcal{L}_{\mathrm{diff}}
    +
    w_{\mathrm{cal}}(k)
    \mathcal{L}_{\mathrm{calib}},
\end{equation}
where $k$ denotes the Stage-II training iteration and $w_{\mathrm{cal}}(k)\in[0,1]$ is a linearly increasing warm-up weight.

\section{Experiments} \label{sec:experiments}
\vspace{-2mm}

\begin{table*}[h]
\centering
\caption{
\textbf{Quantitative comparison on DSEC and MVSEC.}
We compare \alg against open-sourced event-to-depth baselines.
Results marked with $^{*}$ are evaluated after per-image scale-and-shift alignment to the ground-truth LiDAR depth, whereas unmarked results are evaluated directly using metric depth outputs.
Without test-time LiDAR calibration, \alg achieves the best
results across all DSEC metrics and leads in logarithmic errors
and threshold accuracies on MVSEC.
Best and second-best results are shown in \textbf{bold} and
\underline{underline}, respectively, based on unrounded values.}
\vspace{-2mm}
\label{tab:mvsec_dsec_comparison}
\setlength{\tabcolsep}{4pt}
\renewcommand{\arraystretch}{1.12}
\scriptsize

\resizebox{\textwidth}{!}{
\begin{tabular}{llcccccccc}
    \toprule
    & &
    \multicolumn{5}{c}{Error Metrics $\downarrow$} &
    \multicolumn{3}{c}{Accuracy Metrics $\uparrow$} \\
    \cmidrule(lr){3-7}
    \cmidrule(lr){8-10}

    Dataset & Method
    & Abs Rel
    & Sq Rel
    & RMSE
    & RMSE$_{\log}$
    & SILog
    & $\delta < 1.25$
    & $\delta < 1.25^2$
    & $\delta < 1.25^3$ \\
    \midrule

       \multirow{10}{*}{DSEC}
    & E2Depth~\cite{e2depth}
    & 0.344
    & 0.253
    & 13.467
    & 0.376
    & 0.098
    & 0.447
    & 0.755
    & 0.915 \\

    & E2Depth$^{*}$~\cite{e2depth}
    & 0.253
    & 0.130
    & 10.119
    & 0.315
    & 0.107
    & 0.574
    & 0.861
    & 0.956 \\

    & EReFormer~\cite{ereformer}
    & 0.387
    & 0.401
    & 13.954
    & 0.395
    & 0.124
    & 0.486
    & 0.776
    & 0.892 \\

    & EReFormer$^{*}$~\cite{ereformer}
    & 0.286
    & 0.208
    & 11.369
    & 0.325
    & 0.109
    & 0.569
    & 0.839
    & 0.944 \\

    & DepthAnyEvent~\cite{depthanyevent}
    & 0.277
    & 0.170
    & 11.117
    & 0.292
    & 0.051
    & 0.585
    & 0.860
    & 0.955 \\

    & DepthAnyEvent$^{*}$~\cite{depthanyevent}
    & 0.201
    & 0.079
    & 8.880
    & 0.266
    & 0.077
    & 0.664
    & 0.917
    & 0.975 \\

    & DepthAnyEvent-R~\cite{depthanyevent}
    & 0.252
    & 0.128
    & 9.824
    & 0.268
    & 0.045
    & 0.592
    & 0.900
    & 0.971 \\

    & DepthAnyEvent-R$^{*}$~\cite{depthanyevent}
    & 0.191
    & 0.070
    & 8.618
    & 0.244
    & 0.064
    & 0.691
    & 0.930
    & 0.981 \\

    & \textbf{\alg} (ours)
    & \underline{0.149}
    & \underline{0.055}
    & \underline{5.268}
    & \underline{0.203}
    & \underline{0.037}
    & \underline{0.794}
    & \underline{0.949}
    & \underline{0.987} \\ 
    
    & \textbf{\alg-fast} (ours)
    & \textbf{0.146}
    & \textbf{0.055}
    & \textbf{5.046}
    & \textbf{0.196}
    & \textbf{0.034}
    & \textbf{0.810}
    & \textbf{0.953}
    & \textbf{0.987} \\
    \midrule

    \multirow{6}{*}{MVSEC}
    & E2Depth$^{*}$~\cite{e2depth}
    & 0.420
    & 0.806
    & 7.268
    & 0.455
    & 0.213
    & 0.432
    & 0.717
    & 0.868 \\

    & EReFormer$^{*}$~\cite{ereformer}
    & 0.511
    & 1.057
    & 8.373
    & 0.523
    & 0.274
    & 0.391
    & 0.652
    & 0.810 \\

    & DepthAnyEvent$^{*}$~\cite{depthanyevent}
    & \underline{0.373}
    & \underline{0.715}
    & \underline{6.627}
    & 0.449
    & 0.222
    & 0.471
    & 0.747
    & \underline{0.884} \\

    & DepthAnyEvent-R$^{*}$~\cite{depthanyevent}
    & \textbf{0.365}
    & \textbf{0.691}
    & \textbf{6.465}
    & 0.483
    & 0.258
    & \underline{0.489}
    & 0.751
    & 0.878 \\

    & \textbf{\alg} (ours)
    & 0.388
    & 0.993
    & 7.674
    & \textbf{0.423}
    & \textbf{0.167}
    & \textbf{0.521}
    & \textbf{0.768}
    & \textbf{0.885} \\

    & \textbf{\alg-fast} (ours)
    & 0.396
    & 0.900
    & 7.688
    & \underline{0.435}
    & \underline{0.178}
    & 0.459
    & \underline{0.752}
    & 0.880 \\

    \bottomrule
\end{tabular}
}
\end{table*}

\begin{figure}[b]
    \vspace{-3mm}
    \centering
    \includegraphics[width=\columnwidth]{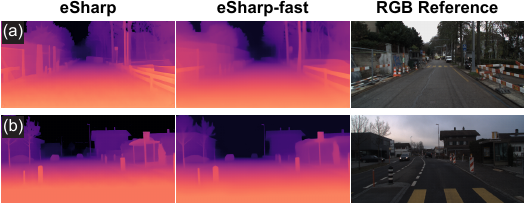}
    \vspace{-6mm}
    \caption{\textbf{\alg \textit{vs.} \alg-fast}. \alg (10-step diffusion) produces sharper depth predictions than \alg-fast (1-step diffusion). The two models produce largely the same image semantics.}
    \label{fig:alg_vs_fast}
    \vspace{-5mm}
\end{figure}

\begin{figure*}[t]
\centering
\includegraphics[width=\textwidth]{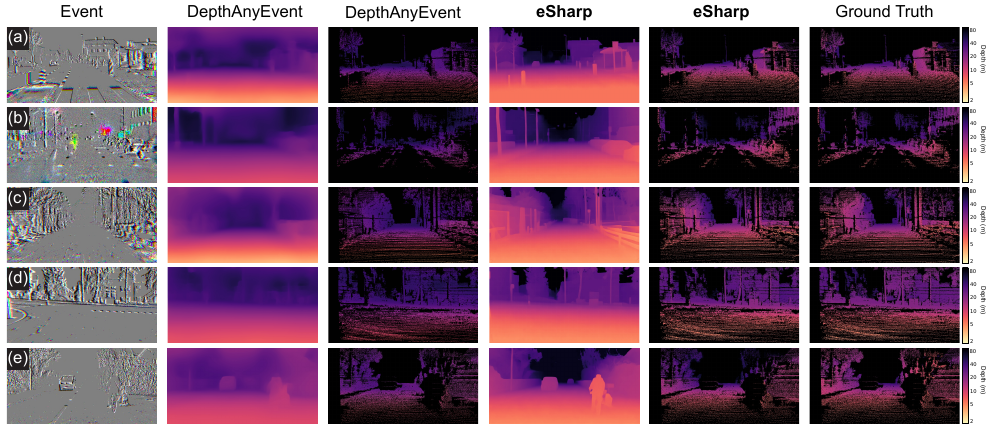}
\vspace{-6mm}
\caption{
\textbf{\alg produces high-fidelity metric depth predictions on DSEC.}
Compared to baseline DepthAnyEvent, \alg produces significantly sharper depth predictions, delineating cars, small obstacles on roadways, and buildings.
$\Omega$ denotes results shown only at valid LiDAR pixels.
All predictions are visualized using the same fixed depth mapping.}
\label{fig:dsec_result}
\vspace{-15pt}
\end{figure*}

\begin{figure*}[t]
\centering
\includegraphics[width=0.8\textwidth]{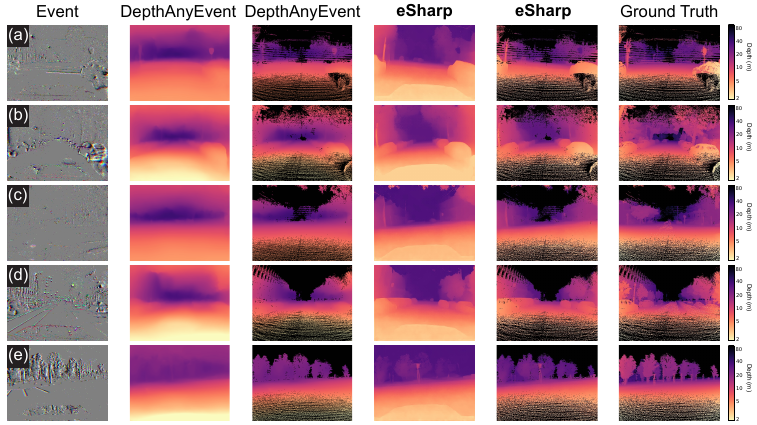}
\vspace{-3mm}
\caption{
\textbf{Metric depth predictions on MVSEC.}
On this dataset with lower-quality imagery (lower resolution and sparser events), \alg still generates sharper, more accurate metric depth predictions than the baseline. 
$\Omega$ denotes results shown only at valid LiDAR pixels.
All predictions are visualized using the same fixed depth mapping.
\alg also reconstructs the vehicle hood (curved region at the bottom).
}
\label{fig:mvsec_result}
\vspace{-4mm}
\end{figure*}

\smallskip \noindent \textbf{Implementation Details.}
Our latent diffusion backbone follows the Marigold~\cite{ke2024repurposing} formulation, and we use DepthAnything V2 Large~\cite{yang2024depth} for pseudo-label generation. Stage I trains the event-conditioned U-Net for $40k$ iterations with learning rate $3\mathrm{e}{-5}$, followed by $16k$ Stage-II iterations jointly optimizing the U-Net and calibration head with learning rates $3\mathrm{e}{-6}$. We use an effective batch size of 8 and train all models on a single NVIDIA A100 80GB GPU. Additional training details are provided in Supplementary \cref{sec:supp_model_details}.

\smallskip \noindent \textbf{Datasets.}
We evaluated \alg on two established event-based monocular depth benchmarks: DSEC~\cite{dsec} and MVSEC~\cite{mvsec}.
DSEC uses a higher-resolution event camera (480 × 640) paired with RGB frames, while MVSEC uses a lower-resolution DAVIS346 event camera (260 × 346) that also provides grayscale intensity frames. We use cropped event representations of $320 \times 640$ for DSEC and $256 \times 344$ for MVSEC. Following standard evaluation protocols used in prior
works~\cite{e2depth,ereformer,depthanyevent},
DSEC contains $19,217$ training samples and $7,167$
validation samples; we train on the MVSEC outdoor day2
sequence and evaluate on outdoor day1 and outdoor night1--3.
MVSEC is comparatively worse in image quality, sample size, and diversity. MVSEC dataset limitations and details of all images from MVSEC and DSEC used in our figures are provided in the Supplementary \cref{sec:supp_dataset}.

MVSEC provides spatially aligned events, intensity frames, and metric depth.
In DSEC, however, the RGB and event cameras have different viewpoints; the temporally corresponding RGB frame was therefore projected into the event-camera coordinate system during pseudo-label generation.
Dense metric pseudo-depth supervision was generated as described in \cref{sec:pseudo_label}.

\smallskip \noindent \textbf{\alg \textit{vs.} \alg-fast.}
With its one-step $x_0$-prediction formulation, \alg-fast produces slightly smoother depth boundaries than \alg, which uses ten-step $v_t$-prediction, as shown in \cref{fig:alg_vs_fast}.
The difference is concentrated primarily around depth discontinuities and leaves the overall scene structure largely unchanged.
Since the aggregate depth metrics are computed only over valid LiDAR pixels, they do not explicitly evaluate the sharpness of the depth map. Hence, the sharper predictions of \alg are not fully reflected in the quantitative metrics.
Nevertheless, \alg-fast runs $6.3\times$ faster than \alg and achieves 14 frames per second (FPS) on DSEC using a single A100 GPU,
details are provided in the Supplementary \cref{sec:supp_model_details}.

\smallskip \noindent \textbf{Baselines and Evaluation Metrics.}
We compared \alg with existing monocular event-to-depth models e2depth~\cite{e2depth}, EReFormer~\cite{ereformer}, DepthAnyEvent~\cite{depthanyevent}, and its recurrent extension DepthAnyEvent-R~\cite{depthanyevent}, all following the same data splits~\cite{depthanyevent}.
\alg predicted metric depth directly and was evaluated without per-frame scale or shift alignment.
For prior methods evaluated as affine-invariant depth estimators, we retained their reported scale-and-shift alignment protocol (denoted as $^\ast$ in \cref{tab:mvsec_dsec_comparison}).
Predictions were evaluated in the event-camera coordinates over a \qtyrange{0}{80}{\metre} depth range.
We report absolute relative error (Abs Rel), squared relative error (Sq Rel), root mean squared error (RMSE), logarithmic RMSE (RMSE$_{\log}$), scale-invariant logarithmic error (SILog), and threshold accuracies $\delta < 1.25$, $\delta < 1.25^2$, $\delta < 1.25^3$.

\smallskip \noindent \textbf{Quantitative Results.}
\Cref{tab:mvsec_dsec_comparison} compares \alg with existing event-based monocular depth baselines on DSEC and MVSEC.
On DSEC, \alg-fast achieved the best error and accuracy metrics, with \alg being the second best.
Notably, both variants directly predict metric depth yet outperform even the relative-depth baselines that use per-image scale-and-shift alignment with ground-truth LiDAR at test time. On MVSEC, all competing baselines use test-time LiDAR alignment. Despite this disadvantage, \alg achieves the best RMSE$_{\log}$ and SILog errors and the best accuracy under all three $\delta$ thresholds.
The aligned baselines retain lower Abs Rel, Sq Rel, and RMSE on MVSEC.

\begin{figure}[h]
    \centering
    \includegraphics[width=\columnwidth]{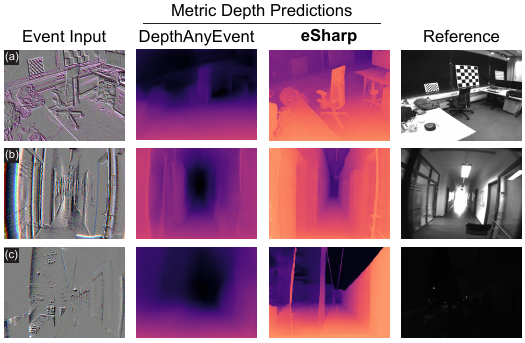}
    \vspace{-6mm}
    \caption{
    \textbf{Zero-shot depth estimation.}
    We evaluated event-to-depth models trained solely on DSEC (driving scenes) on TUM-VIE (diverse egocentric scenes) to qualitatively assess domain transfer.
    \alg still produces reasonable depth estimates, while DepthAnyEvent exhibits weaker transfer skills.
    See grayscale images (different camera pose and field of view) for visual reference.}
    \label{fig:zero_shot}
    \vspace{-2mm}
\end{figure}

\begin{figure}[h]
    \centering
    \includegraphics[width=\columnwidth]{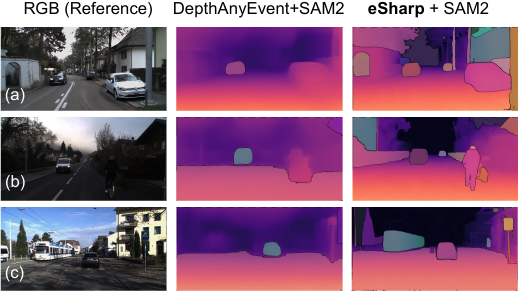}
    \vspace{-6mm}
    \caption{\textbf{\alg predictions enable downstream segmentation.} We qualitatively evaluate SAM2~\cite{ravi2025sam2} directly on dense depth predictions from DepthAnyEvent~\cite{depthanyevent} and \alg. The sharper scene structure preserved by \alg yields more coherent object masks and improved segmentation quality.}
    \label{fig:segmentation}
    \vspace{-4mm}
\end{figure}

\begin{table*}[h]
\centering
\caption{
\textbf{Ablation of two-stage training and metric calibration on \alg.}
Lower is better for error metrics ($\downarrow$), while higher is better
for accuracy metrics ($\uparrow$). The best result for each dataset is
shown in \textbf{bold}.
}
\vspace{-3mm}
\label{tab:calibration_ablation}
\setlength{\tabcolsep}{4pt}
\renewcommand{\arraystretch}{1.12}
\scriptsize

\resizebox{\textwidth}{!}{%
\begin{tabular}{llcccccccc}
    \toprule
    & &
    \multicolumn{5}{c}{Error Metrics $\downarrow$} &
    \multicolumn{3}{c}{Accuracy Metrics $\uparrow$} \\
    \cmidrule(lr){3-7}
    \cmidrule(lr){8-10}

    Dataset & Training Strategy
    & Abs Rel
    & Sq Rel
    & RMSE
    & RMSE$_{\log}$
    & SILog
    & $\delta < 1.25$
    & $\delta < 1.25^2$
    & $\delta < 1.25^3$ \\
    \midrule

    \multirow{3}{*}{DSEC}
    & latent optimization 
    & 0.342
    & 0.189
    & 6.749
    & 0.325
    & 0.039
    & 0.392
    & 0.859
    & 0.972 \\

    & [latent + metric] optimization 
    & 0.239
    & 0.146
    & 6.591
    & 0.285
    & 0.074
    & 0.644
    & 0.890
    & 0.960 \\

    & latent $\rightarrow$ [latent + metric] optimization 
    & \textbf{0.149}
    & \textbf{0.055}
    & \textbf{5.268}
    & \textbf{0.203}
    & \textbf{0.037}
    & \textbf{0.794}
    & \textbf{0.949}
    & \textbf{0.987} \\
    
    \midrule

    \multirow{3}{*}{MVSEC}
    & latent optimization 
    & 0.449
    & 1.167
    & 7.899
    & 0.447
    & \textbf{0.166}
    & 0.486
    & 0.742
    & 0.869 \\

    & [latent + metric] optimization 
    & 0.503
    & 2.035
    & 9.766
    & 0.478
    & 0.191
    & 0.508
    & 0.743
    & 0.858 \\

    & latent $\rightarrow$ [latent + metric] optimization 
    & \textbf{0.388}
    & \textbf{0.993}
    & \textbf{7.674}
    & \textbf{0.423}
    & 0.167
    & \textbf{0.521}
    & \textbf{0.768}
    & \textbf{0.885} \\

    \bottomrule
\end{tabular}%
}
\vspace{-4mm}
\end{table*}

\smallskip \noindent \textbf{Qualitative Results.}
We show metric depth predictions on DSEC
(\cref{fig:dsec_result}) and MVSEC (\cref{fig:mvsec_result}).
\alg recovered sharp depth discontinuities around vehicles, poles, and other structures that produced strong event boundaries, while maintaining coherent depth within objects and surfaces.

\smallskip \noindent \textbf{Ablation Study.}
%
We further study when metric supervision should be introduced during training. \Cref{tab:calibration_ablation} evaluates latent-only optimization, joint latent-and-metric optimization from the start, and our two-stage strategy that introduces metric supervision after latent adaptation. Compared with both latent-only and joint latent-and-metric optimization, our two-stage strategy improves all reported metrics on DSEC and MVSEC, except for a negligible increase in MVSEC SILog ($0.166$ to $0.167$).

\smallskip \noindent \textbf{Zero-Shot Transfer to TUM-VIE.}
To evaluate domain transfer skills of \alg, we take our model checkpoint trained on DSEC and evaluate zero-shot transfer on TUM-VIE~\cite{tumvie}, a dataset that contains diverse scenes - spanning egocentric indoor, low-light, and high-dynamic-range conditions that are far beyond our training distribution (which only contains driving scenes). As shown in \cref{fig:zero_shot}, \alg produces coherent dense depth without dataset-specific adaptation, whereas DepthAnyEvent exhibits incorrect metric scale and weaker object boundaries.

\smallskip \noindent \textbf{Downstream task on eSharp predictions.}
We perform an experiment to validate the sharpness of eSharp's output by applying Segment Anything 2~\cite{ravi2025sam2} to the dense depth map in \cref{fig:segmentation}. The intuition is that if we have accurate depth of objects, objects at different depths can be easily segregated in the depth map - for example, the front and back cars in the first row. eSharp predicts accurate object segments, indicating the sharpness of the predicted depth. In contrast, DepthAnyEvent generates inaccurate object segments (bus in the last row) or misses objects entirely in the second row.

\section{Discussion}
\label{sec:discussion}

Our results show that dense, high-quality supervision is critical for event-based depth estimation. Because our pseudo-label quality depends on sharp frame images, performance is substantially better on DSEC, which provides high-quality frames, than on MVSEC, where lower-quality imagery limits supervision fidelity.
We also utilize frame MDE teacher models because frame cameras perform well under standard conditions, enabling accurate pseudo-label generation. At inference, event cameras retain advantages under challenging conditions such as high dynamic range and fast motion, where frame sensing degrades.
Advances in frame-based MDE models thus can directly improve event-based methods through distillation.
Recent stereo-based foundation models~\cite{wenstereofoundation2025} can also be employed to generate more accurate dense depth labels.
Our findings also motivate the creation of newer datasets with synchronized stereo events, frames, and LiDAR for better dense supervision.

\begin{figure}[t]
    \centering
    \includegraphics[width=\columnwidth]{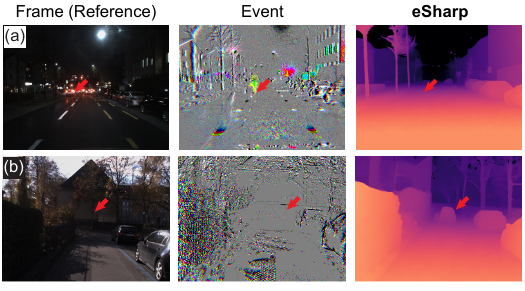}
    \vspace{-6mm}
    \caption{\textbf{Limitations.} \alg metric depth predictions depend on the quality of event streams and can lead to false-positive and false-negative object predictions.}
    \label{fig:limitation}
    \vspace{-6mm}
\end{figure}

Our method also generalizes to out-of-distribution event streams, including indoor scenes, indicating that pretrained diffusion models transfer useful scene and depth priors to the event domain~\cref{fig:zero_shot}. We further use simple temporal binning rather than interpolation or nonuniform binning. Its strong performance suggests that supervision quality and pretrained priors are key contributors.

\smallskip \noindent \textbf{Limitations.}
\alg can miss objects or predict spurious ones when event cues are obscured or ambiguous.
In \cref{fig:limitation}a, strong tail-light responses overwhelm weaker scene edges, causing a missed object.
In \cref{fig:limitation}b, faint lines form a pattern that the driving-scene diffusion prior mistakes for an object.
Training on more diverse scenes, particularly beyond driving, may mitigate these failures.

\vspace{-4mm}

\section{Conclusion}
\label{sec:conclusion}

\vspace{-1mm}

We introduced a diffusion model for sharp monocular metric depth estimation from event streams by adapting pretrained diffusion models with dense pseudo-label supervision. Our framework includes two models for different speed requirements: (a) \alg, a multi-step diffusion model that produces high-fidelity depth maps with fine spatial detail, and (b) \alg-fast, a one-step diffusion model that is approximately $6\times$ faster with a small reduction in fine-detail quality.
Our method substantially improves both quantitative accuracy and prediction quality over prior monocular event-based approaches, while directly estimating metric depth. On the higher-resolution DSEC dataset, it also outperforms the scale-relative depth predictions of state-of-the-art methods, which are scale-aligned using LiDAR maps. More broadly, our results demonstrate that pretrained diffusion models can be effectively adapted to event-based depth prediction even with limited high-quality training data.

\section*{Acknowledgments}
\label{sec:acknowledgments}

This work is supported by the National Institutes of Health (NIH) under award number R01DE032051 (DeepDOF). 
We thank Kushal Vyas for his helpful discussions and suggestions throughout this work.

{
  \small
  \bibliographystyle{ieeenat_fullname}
  \bibliography{references}
}

\clearpage
\onecolumn
\begin{center}
  {\LARGE\bfseries Supplementary Material}
\end{center}
\vspace{1em}
\section{Pseudo-Label Generation Details}
\label{sec:supp_pseudo_label_generation}

\paragraph{Dense accurate depth map generation.}

The camera-matrix homography used by \cite{depthanyevent,pan2024srfnet,gehrig2024low} provides an efficient, calibration-only transformation between the RGB and event-camera viewpoints, without requiring scene geometry at inference time. However, as noted by Gehrig et al.~\cite{gehrig2024low}, this transformation does not provide exact pixel-wise correspondence, since a single global homography cannot fully account for depth-dependent parallax. Nevertheless, it offers a strong and computationally inexpensive alignment baseline. We therefore investigate whether incorporating monocular depth can refine this coarse geometric warping from the frame-camera coordinate system to the event-camera coordinate system and produce more accurate event-space supervision.

We compare this baseline with several depth-aware warping configurations based on state-of-the-art monocular depth estimation (MDE) models Depth Anything v2~\cite{yang2024depth} and Depth Anything v3~\cite{lin2025depth}. The first column of \cref{tab:pseudo_label_depth_comparison} specifies the method used to warp the original RGB frame into the event-camera image plane. The second column specifies the model used to estimate depth from the resulting event-space RGB image. ``DA2/LiDAR'' denotes Depth Anything v2 followed by robust metric alignment to the sparse event-view LiDAR, whereas
``DA3/LiDAR'' denotes Depth Anything v3 depth, followed by an additional robust scale-shift alignment to LiDAR. For the depth-aware warps, the source-view depth is likewise aligned to LiDAR reprojected into the RGB camera before performing the geometric projection. The 10\,m baseline instead assumes a single plane for the entire scene at 10 m. The ``DA3-metric'' baseline directly uses the Depth Anything v3 metric depth without LiDAR-based alignment.

We evaluate all configurations on the same set of 1,000 frames, uniformly sampled without replacement from the DSEC frames across all 41  in \cref{tab:pseudo_label_depth_comparison}. Errors are calculated with respect to the event-view LiDAR depth only for pixels that contain valid LiDAR measurements and valid predictions. Depths are restricted to the range $[0,80]$\,m. For this ablation, pixels for which either the prediction or LiDAR depth reaches the 80\,m upper bound are excluded. Because LiDAR is available during pseudo-label generation, the same sparse measurements are used for metric alignment and evaluation. The experiment, therefore, measures the attainable quality of LiDAR-calibrated dense supervision rather than the generalization to held-out monocular depth.

\begin{table}[h]
\centering
\caption{DSEC event-space pseudo-label depth accuracy on 1000 samples after clipping depth to 0.5--80~m. Lower is better for MAE, RMSE, and AbsRel; higher is better for $\delta$. DA stands for Depth Anything.}
\label{tab:pseudo_label_depth_comparison}
\begin{tabular}{llrrrrrr}
\toprule
Warping method&Image-to-depth method& MAE $\downarrow$ & RMSE $\downarrow$ & AbsRel $\downarrow$ & $\delta_1$ $\uparrow$ & $\delta_2$ $\uparrow$ & $\delta_3$ $\uparrow$ \\
\midrule
Camera-matrix homography & DA2/LiDAR & 3.033& 5.683& 0.120& \uline{0.870}& \uline{0.963}& 0.986\\
10 m plane&DA2/LiDAR & 3.057& 5.721& \uline{0.120}& 0.868& 0.963& \uline{0.986}\\
DA3-LiDAR warp&DA3/LiDAR & \textbf{2.941}& \uline{5.642}& 0.135& 0.861& 0.956& 0.982\\
DA3-metric warp & DA3-metric estimation & 7.669& 9.639& 0.359& 0.244& 0.917& 0.977\\
DA2-LiDAR warp & DA2/LiDAR & \uline{2.983}& \textbf{5.572}& \textbf{0.117}& \textbf{0.873}& \textbf{0.966}& \textbf{0.987}\\
\bottomrule
\end{tabular}
\end{table}

\paragraph{Metric alignment for pseudo labels.}
As described in \cref{sec:pseudo_label}, $\mathbf{d}_{\mathrm{rel}}$ lacks metric scale, so for DSEC we estimate a per-frame affine mapping $(a^\ast,b^\ast)$ to the available sparse ground-truth LiDAR measurements.
We observed that we achieve better pseudo-label alignment by fitting in disparity space rather than depth space.
We do the alignment in disparity space $g$. Let $g_{\mathrm{rel}}$ denote the inverse depth (disparity) predicted by the Depth Anything V2 MDE model, $g_{\mathrm{abs}}^\ast$ denote the inverse depth from the LiDAR ground truth over valid pixels, and $\Omega$ denote the set of pixels with valid LiDAR measurements; the affine parameters are obtained using least-squares fitting over valid pixels,

\begin{equation}
(a^\ast,b^\ast)
=
\arg\min_{a,b}
\sum_{u\in\Omega}
\left(
a g_{\mathrm{rel}}(u)+b-g_{\mathrm{abs}}^\ast(u)
\right)^2.
\end{equation}

The resulting dense aligned inverse depth is $\tilde{g}_{\mathrm{abs}} = a^\ast{g}_{\mathrm{rel}} + b^\ast$; for the DSEC dataset, we then convert this result to metric depth $\tilde{\mathbf{d}}$ using the provided camera calibration. For the MVSEC dataset, the low-resolution grayscale intensity observations (provided by the event camera itself) provide limited appearance information to the image-based depth teacher, so we apply quantile matching during pseudo-label generation to improve robustness.

\paragraph{Pseudo-Label Alignment on MVSEC.}
As discussed in the main paper, MVSEC provides low-resolution grayscale intensity observations
rather than the richer RGB inputs available in DSEC. These observations provide limited appearance cues for the image-based depth teacher, making
pseudo-depth generation more susceptible to large prediction errors. We
therefore apply quantile matching during MVSEC pseudo-label generation for better metric alignment and
evaluate its effect in Table~\ref{tab:mvsec_pseudo_alignment}.

Let $g_{\mathrm{rel}}(u)$ denote the relative inverse-depth prediction (disparity) from Depth Anything V2 and $g_{\mathrm{abs}}^\ast(u)$ the corresponding inverse metric depth at valid pixels $u\in\Omega$. We sample \(K=64\) quantile levels \(q_i\) uniformly over the 5th-95th percentile range and form matched knots
\begin{equation}
\xi_i = Q_{g_{\mathrm{rel}}(\Omega)}(q_i), \qquad
\eta_i = Q_{g_{\mathrm{abs}}^\ast(\Omega)}(q_i),
\qquad i=0,\ldots,K-1,
\end{equation}

where \(Q_x(q)\) denotes the \(q\)-quantile of \(x\). A monotonic piecewise-linear mapping \(f\) is constructed from the matched pairs \((\xi_i,\eta_i)\) and applied to the dense teacher prediction. Values outside the fitted quantile range are clamped to the nearest endpoint. The aligned inverse depth is \(\tilde g(u)=f(g_{\mathrm{rel}}(u))\), which is subsequently converted to the dense metric pseudo-label \(\tilde{\mathbf d}\).

In \cref{tab:mvsec_pseudo_alignment}, we compare the affine least-squares (LS) alignment applied in depth, inverse-depth, and log-depth spaces, with quantile matching. Quantile matching provides the strongest overall pseudo-label quality on MVSEC. On DSEC, quantile matching provides only marginal changes relative to the standard LS alignment, so we retain the simpler least-squares procedure.

\begin{table}[h]
    \centering
    \caption{
    \textbf{Pseudo-label alignment quality on MVSEC.}
    We compare the raw teacher prediction, affine least-squares (LS) alignment
    in different spaces, and quantile matching. Metrics are computed against
    the available metric depth. Lower is better for error metrics
    ($\downarrow$), while higher is better for accuracy metrics ($\uparrow$).
    The best result for each metric is shown in bold.
    }
    \label{tab:mvsec_pseudo_alignment}

    \setlength{\tabcolsep}{4.5pt}
    \renewcommand{\arraystretch}{1.12}
    \small

    \begin{tabular}{lcccccc}
        \toprule
        Method
        & Abs Rel $\downarrow$
        & Sq Rel $\downarrow$
        & RMSE $\downarrow$
        & $\delta_1$ $\uparrow$
        & $\delta_2$ $\uparrow$
        & $\delta_3$ $\uparrow$ \\
        \midrule

        Raw prediction
        & 0.306
        & \textbf{0.225}
        & 9.641
        & 0.463
        & 0.727
        & 0.853 \\

        Depth affine LS
        & 0.407
        & 0.784
        & 7.196
        & 0.452
        & 0.762
        & 0.877 \\

        Inverse-depth affine LS
        & 0.302
        & 0.328
        & 8.517
        & 0.498
        & 0.777
        & 0.891 \\

        Log-depth affine LS
        & 0.315
        & 0.458
        & 7.521
        & 0.529
        & 0.779
        & 0.902 \\

        Quantile matching
        & \textbf{0.268}
        & 0.591
        & \textbf{6.499}
        & \textbf{0.652}
        & \textbf{0.859}
        & \textbf{0.935} \\

        \bottomrule
    \end{tabular}%
\end{table}

\section{Additional Model Details}
\label{sec:supp_model_details}

We provide additional implementation details for \alg and \alg-fast, including their trainable parameter counts, optimization settings, training-strategy ablations, and inference efficiency.

\paragraph{Trainable Parameters.}
As described in the main paper, the first stage adapts the pretrained diffusion backbone to event inputs by fine-tuning the event-conditioned U-Net, while the VAE and text encoder remain frozen. \alg and \alg-fast contain 865.9M and 867.6M trainable U-Net parameters, respectively. In the subsequent metric calibration stage, both models additionally optimize the same lightweight calibration head, which introduces only 1.9K trainable parameters. The corresponding parameter counts are summarized in \cref{tab:supp_params}.

As shown in \cref{tab:calibration_ablation} in the main paper, the second stage of training involves optimizing [latent + metric] jointly.
For each dataset variant, DSEC and MVSEC, the models use the same architecture and therefore have identical parameter counts.

\begin{table}[h]
    \centering
    \caption{
    \textbf{Trainable parameter counts for latent optimization and metric optimization stages.}
    The metric optimization stage introduces only a lightweight calibration head in addition to the
    event-conditioned U-Net.
    }
    \label{tab:supp_params}

    \setlength{\tabcolsep}{6pt}
    \renewcommand{\arraystretch}{1.12}
    \small
    \begin{tabular}{llccc}
        \toprule
        Method & Stage & Trainable modules
        & Trainable params. & Frozen params. \\
        \midrule

        \multirow{2}{*}{\alg}
        & latent optimization
        & U-Net
        & 865.9M
        & 424.0M \\

        & [latent + metric] optimization
        & U-Net + calibration head
        & 865.9M + 1.9K
        & 424.0M \\

        \midrule

        \multirow{2}{*}{\alg-fast}
        & latent optimization
        & U-Net
        & 867.6M
        & 424.0M \\

        & [latent + metric] optimization
        & U-Net + calibration head
        & 867.6M + 1.9K
        & 424.0M \\
        \bottomrule
    \end{tabular}
\end{table}

\FloatBarrier

\rd{ Pick additional model details to present in Supplementary: this is mentioned in main paper section 5: implementation detail:}

\paragraph{\alg Training Details.}
We optimize \alg using Adam with an IterExponential learning rate scheduler, with a final learning rate (lr) ratio of $0.02$ and 200 warm-up iterations.
The same optimizer and scheduler configuration is used for both datasets and both training stages.
We set $\lambda_{\mathrm{aff}}=0.7, \lambda_{\mathrm{metric}}=0.3$
(details of which are shown in the ablation study in \cref{tab:calibration_weight_ablation}), with equal weighting of the scale and shift regression terms. 
Loss coefficient $w_{\mathrm{cal}}(k)$ linearly ramps up from 0 to 1 during the first 2k iterations of the [latent+metric] optimization.
\ad{\sout{For validation and final inference, we use 10 denoising steps with $v_t$-prediction for \alg, one denoising step with $x_0$-prediction for \alg-fast.}}
For validation and final inference, we use 10 denoising steps with $v_t$-prediction for \alg.

\paragraph{\alg-fast Training Details.}
Training \alg-fast follows the same overall framework as \alg, with modifications to the latent-optimization and metric-calibration procedures. During latent optimization, we follow the Lotus training protocol and optimize the one-step $x_0$-prediction model using AdamW with a learning rate of $3\times10^{-6}$, weight decay of $0.01$, $(\beta_1,\beta_2)=(0.9,0.999)$, and 200 warm-up iterations under a constant learning-rate schedule. We use $\epsilon=10^{-8}$ and clip the gradient norm to $1.0$. Before joint latent-and-metric optimization, we briefly freeze the U-Net and optimize only the calibration head to obtain a stable initialization. We then unfreeze the U-Net and jointly optimize the backbone and calibration head using Adam, with the same calibration architecture and losses as \alg.
For validation and final inference, we use one denoising step with $x_0$-prediction for \alg-fast.

\paragraph{\alg-fast Training Strategy Ablation.}
Similar to \cref{tab:calibration_ablation} in the main paper, \cref{tab:calibration_ablation_fast} below reports the corresponding training-strategy ablation for \alg-fast, comparing latent-only optimization, joint latent-and-metric optimization from the start, and our sequential latent $\rightarrow$ [latent + metric] optimization on DSEC and MVSEC.
Overall, our sequential latent $\rightarrow$ [latent + metric] optimization
achieves the best performance on most metrics across both datasets.

\begin{table*}[h]
\centering
\caption{
\textbf{Ablation of two-stage training and metric calibration on \alg-fast.}
Lower is better for error metrics ($\downarrow$), while higher is better
for accuracy metrics ($\uparrow$). The best result for each dataset is
shown in \textbf{bold}.
}
\vspace{-4mm}
\label{tab:calibration_ablation_fast}
\setlength{\tabcolsep}{4pt}
\renewcommand{\arraystretch}{1.12}
\scriptsize

\resizebox{\textwidth}{!}{%
\begin{tabular}{llcccccccc}
    \toprule
    & &
    \multicolumn{5}{c}{Error Metrics $\downarrow$} &
    \multicolumn{3}{c}{Accuracy Metrics $\uparrow$} \\
    \cmidrule(lr){3-7}
    \cmidrule(lr){8-10}

    Dataset & Training Strategy
    & Abs Rel
    & Sq Rel
    & RMSE
    & RMSE$_{\log}$
    & SILog
    & $\delta < 1.25$
    & $\delta < 1.25^2$
    & $\delta < 1.25^3$ \\
    \midrule

    \multirow{3}{*}{DSEC}
    & latent optimization
    & 0.176
    & 0.058
    & 5.792
    & 0.232
    & 0.045
    & 0.720
    & \textbf{0.956}
    & \textbf{0.988} \\

    & [latent + metric] optimization
    & 0.271
    & 0.108
    & 6.858
    & 0.371
    & 0.068
    & 0.391
    & 0.749
    & 0.927 \\

    & latent $\rightarrow$ [latent + metric] optimization
    & \textbf{0.146}
    & \textbf{0.055}
    & \textbf{5.046}
    & \textbf{0.196}
    & \textbf{0.034}
    & \textbf{0.810}
    & 0.953
    & 0.987 \\

    \midrule

    \multirow{3}{*}{MVSEC}
    & latent optimization
    & 0.399
    & \textbf{0.824}
    & 7.693
    & 0.457
    & 0.191
    & 0.392
    & 0.714
    & 0.872 \\

    & [latent + metric] optimization
    & 0.480
    & 1.073
    & \textbf{7.519}
    & 0.470
    & 0.185
    & 0.409
    & 0.698
    & 0.848 \\

    & latent $\rightarrow$ [latent + metric] optimization
    & \textbf{0.396}
    & 0.900
    & 7.688
    & \textbf{0.435}
    & \textbf{0.178}
    & \textbf{0.459}
    & \textbf{0.752}
    & \textbf{0.880} \\

    \bottomrule
\end{tabular}%
}
\vspace{-4mm}
\end{table*}

\paragraph{Smooth L1 Affine Regression.}
In the metric optimization stage, the predicted affine scale and shift are supervised using a
Smooth L1 loss against the least-squares targets described in the main paper.
Compared with a standard $\ell_2$ objective, Smooth L1 behaves quadratically
for small residuals and linearly for larger residuals, providing stable
optimization while reducing sensitivity to occasional large errors in the
estimated affine targets. The affine regression loss is defined as
\begin{equation}
    \mathcal{L}_{\mathrm{aff}}
    =
    \frac{1}{2}
    \left[
        \rho(s-s^\ast)
        +
        \rho(\beta-\beta^\ast)
    \right],
\end{equation}
where $\rho(\cdot)$ denotes the Smooth L1 function and
$(s^\ast,\beta^\ast)$ are the per-sample least-squares scale and shift targets.
We weight the scale and shift terms equally in all experiments.

\paragraph{Inference Efficiency.}
For \alg and \alg-fast, all reported results were obtained without test-time ensembling.
As reported in \cref{sec:experiments}, \alg-fast runs $6.3\times$ faster than
\alg and achieves approximately 14 frames per second (FPS) on DSEC using a
single A100 GPU. \Cref{tab:efficiency} provides the corresponding efficiency
details and compares our diffusion-based models with
DepthAnyEvent~\cite{depthanyevent} in terms of mean inference time per frame
and peak allocated GPU memory under the DSEC and MVSEC evaluation settings.
Although iterative diffusion inference incurs a higher computational cost than
the feed-forward baseline, \alg-fast substantially reduces this overhead.

\begin{table}[h]
    \centering
    \caption{
    \textbf{Inference efficiency on DSEC and MVSEC.}
    Spatial (H$\times$W) resolution of binned event data for each dataset is shown here.
    We report mean inference latency, throughput, and peak allocated GPU memory.
    Lower is better for latency and VRAM ($\downarrow$); higher is better for frames per second (FPS) ($\uparrow$).}
    \label{tab:efficiency}

    \setlength{\tabcolsep}{6pt}
    \renewcommand{\arraystretch}{1.12}
    \small

    \begin{tabular}{lllccc}
        \toprule
        Dataset & Resolution & Method
        & Latency (\unit{\second\per\frameunit}) $\downarrow$
        & FPS $\uparrow$
        & VRAM (GB) $\downarrow$ \\
        \midrule
    
        \multirow{3}{*}{DSEC}
        & \multirow{3}{*}{$320\times640$}
        & Depth AnyEvent~\cite{depthanyevent}
        & 0.012
        & 83
        & 0.29 \\
        &
        & \textbf{\alg}
        & 0.47
        & 2.1
        & 5.6 \\
        &
        & \textbf{\alg-fast}
        & 0.073
        & 14
        & 2.3 \\
    
        \midrule
    
        \multirow{3}{*}{MVSEC}
        & \multirow{3}{*}{$256\times344$}
        & Depth AnyEvent~\cite{depthanyevent}
        & 0.0097
        & 100
        & 0.15 \\
        &
        & \textbf{\alg}
        & 0.36
        & 2.8
        & 5.2 \\
        &
        & \textbf{\alg-fast}
        & 0.049
        & 20.4
        & 2.0 \\
    
        \bottomrule
    \end{tabular}

\end{table}

\FloatBarrier

\section{Metric Depth Calibration}
\label{sec:supp_metric_depth_calibration}

We provide further details and ablations for calibrating both the RGB-derived pseudo-labels and the event-to-depth predictions described in the main paper.

\noindent\textbf{Event-to-Depth Prediction Calibration.}
To directly supervise the calibration head, we compute the
scale $s$ and shift $\beta$ that best align the current decoded
prediction $\hat{d}$ with the dense metric pseudo-depth
$\tilde{d}$. Let $\mathcal{P}$ denote the full image domain,
containing all $H \times W$ pixels. The target affine
parameters are obtained by least-squares fitting:

\begin{equation}
(s^*, \beta^*) =
\arg\min_{s,\beta}
\sum_{u \in \mathcal{P}}
\left(
s\hat{d}(u) + \beta - \tilde{d}(u)
\right)^2.
\label{eq:calibration_target}
\end{equation}

Unlike the sparse LiDAR-valid pixel set $\Omega$ used
during pseudo-label generation, calibration supervision
uses the dense metric pseudo-depth over the entire image.
The resulting $(s^*, \beta^*)$ serve as frame-specific
regression targets for the affine parameters predicted
by the calibration head.

\paragraph{Calibration-Loss Weighting.}
We further study the relative contributions of the affine and metric-depth supervision terms in \cref{eq:calib} by varying $\lambda_{\mathrm{aff}}$ and $\lambda_{\mathrm{metric}}$ in \cref{tab:calibration_weight_ablation}. For \alg, combining the two objectives improves overall performance, with $\lambda_{\mathrm{aff}}=0.7$ and $\lambda_{\mathrm{metric}}=0.3$ providing the strongest results. For \alg-fast, the metric-only ($0/1$), balanced ($0.5/0.5$), and $0.7/0.3$ configurations perform comparably, with only marginal differences across metrics. We therefore use $\lambda_{\mathrm{aff}}=0.7$ and $\lambda_{\mathrm{metric}}=0.3$ for both models in the main paper for consistency.

\begin{table*}[h]
    \centering
    \caption{
    \textbf{Ablation of calibration-loss weighting on DSEC.}
    $\lambda_{\mathrm{aff}}$ and $\lambda_{\mathrm{metric}}$ control the
    affine-parameter and metric-depth supervision, respectively.
    Lower is better for error metrics ($\downarrow$), while higher is better
    for accuracy metrics ($\uparrow$).
    Best and second-best results within each model are shown in
    \textbf{bold} and \underline{underline}, respectively.
    }
    \label{tab:calibration_weight_ablation}

    \setlength{\tabcolsep}{4pt}
    \renewcommand{\arraystretch}{1.12}
    \scriptsize

    \resizebox{\textwidth}{!}{%
    \begin{tabular}{lllcccccccc}
        \toprule
        & & &
        \multicolumn{5}{c}{Error Metrics $\downarrow$} &
        \multicolumn{3}{c}{Accuracy Metrics $\uparrow$} \\
        \cmidrule(lr){4-8}
        \cmidrule(lr){9-11}

        Method
        & $\lambda_{\mathrm{aff}}$
        & $\lambda_{\mathrm{metric}}$
        & Abs Rel
        & Sq Rel
        & RMSE
        & RMSE$_{\log}$
        & SILog
        & $\delta < 1.25$
        & $\delta < 1.25^2$
        & $\delta < 1.25^3$ \\
        \midrule

         \multirow{4}{*}{\alg}
        & 0.0 & 1.0
        & 0.262
        & 0.132
        & 5.973
        & 0.266
        & \textbf{0.035}
        & 0.598
        & 0.912
        & 0.976 \\

        & 0.3 & 0.7
        & \underline{0.161}
        & \underline{0.066}
        & \underline{5.341}
        & \underline{0.212}
        & 0.039
        & \underline{0.782}
        & \underline{0.944}
        & \underline{0.984} \\

        & 0.5 & 0.5
        & 0.166
        & 0.072
        & 5.390
        & 0.214
        & 0.038
        & 0.777
        & 0.940
        & 0.983 \\

        & 0.7 & 0.3
        & \textbf{0.149}
        & \textbf{0.055}
        & \textbf{5.268}
        & \textbf{0.203}
        & \underline{0.037}
        & \textbf{0.794}
        & \textbf{0.949}
        & \textbf{0.987} \\
        \midrule

        \multirow{5}{*}{\alg-fast}
        & 0.0 & 1.0
        & \underline{0.146}
        & \textbf{0.055}
        & \underline{5.031}
        & \underline{0.196}
        & \textbf{0.034}
        & 0.809
        & \textbf{0.953}
        & \textbf{0.987} \\

        & 0.3 & 0.7
        & 0.149
        & \underline{0.056}
        & 5.085
        & 0.200
        & \underline{0.035}
        & 0.797
        & \underline{0.951}
        & \textbf{0.987} \\

        & 0.5 & 0.5
        & \textbf{0.145}
        & \textbf{0.055}
        & \textbf{5.025}
        & \textbf{0.195}
        & \textbf{0.034}
        & \textbf{0.813}
        & \textbf{0.953}
        & \textbf{0.987} \\

        & 0.7 & 0.3
        & \underline{0.146}
        & \textbf{0.055}
        & 5.046
        & \underline{0.196}
        & \textbf{0.034}
        & \underline{0.810}
        & \textbf{0.953}
        & \textbf{0.987} \\

        & 1.0 & 0.0
        & 0.224
        & 0.133
        & 6.544
        & 0.266
        & 0.062
        & 0.687
        & 0.904
        & \underline{0.964} \\

        \bottomrule
    \end{tabular}%
    }
\end{table*}

\FloatBarrier

\section{Dataset}
\label{sec:supp_dataset}

\paragraph{Limitations of the MVSEC dataset.}
There is limited scene variation in MVSEC. There are a total of five sequences: \texttt{outdoor\_day1}, \texttt{outdoor\_day2},
\texttt{outdoor\_night1}, \texttt{outdoor\_night2}, and
\texttt{outdoor\_night3}, and only part of the \texttt{outdoor\_day2} is used for training. Since the frame images also come from the low-resolution event camera, their quality is often poor.

\paragraph{Details of samples used for visualization}

In \cref{tab:visualization_sample_details} below, we list all filenames and datasets used in the visualization figures for reproducibility; none of them are part of the training split.

\begin{table}[h]
    \centering
    \caption{Sample details used to generate the qualitative results shown in the main paper.}
    \label{tab:visualization_sample_details}
    \small
    \begin{tabular}{@{}c l@{}}
        \toprule
        \textbf{Figure} & \textbf{Dataset and sample identifier} \\
        \midrule

        \multirow{3}{*}{Fig.~1}
        & \texttt{\detokenize{dsec_interlaken_00_b_rgb_000950}} \\
        & \texttt{\detokenize{dsec_interlaken_01_a_rgb_000090}} \\
        & \texttt{\detokenize{dsec_zurich_city_09_d_0000000094}} \\
        \addlinespace[2pt]

        \midrule

        Fig.~2
        & \texttt{\detokenize{dsec_interlaken_00_f_0000000020}} \\
        \addlinespace[2pt]

        \midrule

        Fig.~3
        & \texttt{\detokenize{dsec_pipeline_interlaken_00_f_0000000720}} \\
        \addlinespace[2pt]

        \midrule

        \multirow{2}{*}{Fig.~4}
        & \texttt{\detokenize{dsec_zurich_city_05_b_0000000043}} \\
        & \texttt{\detokenize{dsec_pipeline_interlaken_00_f_0000000720}} \\
        \addlinespace[2pt]

        \midrule

        \multirow{5}{*}{Fig.~5}
        & \texttt{\detokenize{dsec_pipeline_interlaken_00_f_0000000720}} \\
        & \texttt{\detokenize{dsec_zurich_city_09_d_0000000321}} \\
        & \texttt{\detokenize{dsec_zurich_city_05_b_0000000043}} \\
        & \texttt{\detokenize{dsec_zurich_city_05_a_0000000254}} \\
        & \texttt{\detokenize{dsec_interlaken_00_g_0000000217}} \\
        \addlinespace[2pt]

        \midrule

        \multirow{5}{*}{Fig.~6}
        & \texttt{\detokenize{mvsec_outdoor_night3_0000005331}} \\
        & \texttt{\detokenize{mvsec_outdoor_night3_0000003977}} \\
        & \texttt{\detokenize{mvsec_outdoor_night3_0000002605}} \\
        & \texttt{\detokenize{mvsec_outdoor_night1_0000001352}} \\
        & \texttt{\detokenize{mvsec_outdoor_day1_0000001362}} \\
        \addlinespace[2pt]

        \midrule

        \multirow{3}{*}{Fig.~7}
        & \texttt{\detokenize{TUM-VIE_running-easy_01316}} \\
        & \texttt{\detokenize{TUM-VIE_running-easy_00536}} \\
        & \texttt{\detokenize{TUM-VIE_floor2-dark_01117}} \\
        \addlinespace[2pt]

        \midrule

        \multirow{3}{*}{Fig.~8}
        & \texttt{\detokenize{dsec_zurich_city_05_a_0000000768}} \\
        & \texttt{\detokenize{dsec_interlaken_00_g_0000000217}} \\
        & \texttt{\detokenize{dsec_zurich_city_07_a_0000000135}} \\
        \addlinespace[2pt]

        \midrule

        \multirow{2}{*}{Fig.~9}
        & \texttt{\detokenize{dsec_zurich_city_09_d_0000000321}} \\
        & \texttt{\detokenize{dsec_zurich_city_06_a_0000000756}} \\

        \bottomrule
    \end{tabular}
\end{table}

\end{document}